\def\year{2022}\relax
\documentclass[letterpaper]{article} % DO NOT CHANGE THIS
\usepackage{aaai22}  % DO NOT CHANGE THIS
\usepackage{times}  % DO NOT CHANGE THIS
\usepackage{helvet}  % DO NOT CHANGE THIS
\usepackage{courier}  % DO NOT CHANGE THIS
\usepackage[hyphens]{url}  % DO NOT CHANGE THIS
\usepackage{graphicx} % DO NOT CHANGE THIS
\usepackage{natbib}  % DO NOT CHANGE THIS AND DO NOT ADD ANY OPTIONS TO IT
\usepackage{caption} % DO NOT CHANGE THIS AND DO NOT ADD ANY OPTIONS TO IT
\DeclareCaptionStyle{ruled}{labelfont=normalfont,labelsep=colon,strut=off} % DO NOT CHANGE THIS
\usepackage{algorithm}
\usepackage{newfloat}
\usepackage{listings}
\floatstyle{ruled}
\newfloat{listing}{tb}{lst}{}
\floatname{listing}{Listing}

\title{Interpretable Local Tree Surrogate Policies}
\author {
    John Mern,\textsuperscript{\rm 1}
    Sidhart Krishnan, \textsuperscript{\rm 2}
    Anil Yildiz, \textsuperscript{\rm 1}
    Kyle Hatch, \textsuperscript{\rm 2}
    Mykel J. Kochenderfer \textsuperscript{\rm 1}
}
\affiliations {
    \textsuperscript{\rm 1} Department of Aeronautics and Astronautics, Stanford University\\
    \textsuperscript{\rm 2} Department of Computer Science, Stanford University \\
}
\usepackage{amsmath,amssymb, bbm}
\usepackage[noend]{algpseudocode}
\usepackage{booktabs}
\usepackage[nameinlink]{cleveref}
\usepackage{subcaption}
\usepackage{multirow}

\usepackage{flushend}

\begin{document}

\maketitle
% 7 + 2 Pages
\begin{abstract}
% State the problem, your approach and solution, and the main contributions of the paper. Include little if any background and motivation. Be factual but comprehensive. The material in the abstract should not be repeated later word for word in the paper. 
High-dimensional policies, such as those represented by neural networks, cannot be reasonably interpreted by humans. 
This lack of interpretability reduces the trust users have in policy behavior, limiting their use to low-impact tasks such as video games.
Unfortunately, many methods rely on neural network representations for effective learning. 
In this work, we propose a method to build predictable policy trees as surrogates for policies such as neural networks.
The policy trees are easily human interpretable and provide quantitative predictions of future behavior. 
We demonstrate the performance of this approach on several simulated tasks. 
\end{abstract}
\section{Introduction}
% What is the problem?
% Why is it interesting and important?
% Why is it hard? (E.g., why do naive approaches fail?)
% Why hasn't it been solved before? (Or, what's wrong with previous proposed solutions? How does mine differ?)
% What are the key components of my approach and results? Also include any specific limitations.

Deep reinforcement learning has achieved state of the art performance in several challenging task domains~\cite{mnih2015}. 
Much of that performance comes from the use of highly expressive neural networks to represent policies.
Humans are generally unable to meaningfully interpret neural network parameters, which has lead to the common view of neural networks as ``black-box" functions. 
Poor interpretability often leads to a lack of trust in neural networks and use of other more transparent, though potentially less high-performing policy models. 
This is often referred to as the performance-transparency trade-off. % in machine learning.

% The importance users place on interpretability often varies by the task performed.
Interpretability is important for high-consequence tasks.
Domains in which neural networks have already been applied, such as image classification, often do not require interpretable decisions because the consequences of mislabeling images are typically low. 
In many potential applications of deep reinforcement learning, such as autonomous vehicle control, erroneous actions may be costly and dangerous.
In these cases, greater trust in policy decisions is typically desired before systems are deployed.

% Our work is motivated by tasks in which a policy is enacted with direct cooperation of a human who must approve of policy actions or directly enact policy recommendations. 
Our work is motivated by tasks in which a human interacts directly with the policy, either by approving agent actions before they are taken or by enacting recommendations directly. 
This is often referred to having a human ``in the loop".
An example is an automated cyber security incident response system that provides recommendations to a human analyst.
% An example of this is an cyber awareness agent that provides recommended courses of action to a human security analyst. 
In these cases, knowing the extended course of actions before committing to the recommendation can enhance the trust of the human operator.
%TODO COVID Policies comment

It is difficult for humans to holistically interpret models with even a small number of interacting terms~\cite{lipton2018}. 
% Instead, models are typically interpreted in only a limited scope. 
% Global model interpretations are those that are valid across the complete input space of the model. 
% Linear regression models are globally interpretable, as the feature coefficients encode transparent relationships that hold across the entire input domain. 
% Local interpretations are those that only hold for specific inputs or groups of inputs and do not necessarily generalize.
% Neural networks are not globally transparent due to their high number of parameters and non-linear interactions. 
Neural networks commonly have several thousand parameters and non-linear interactions, making holistic interpretation infeasible.
Some existing methods constrain neural networks architectures and train them to learn human interpretable features during task learning. % and are intrinsically transparent.
The training and architecture constraints, however, can degrade performance compared to an unconstrained policy. 
A common approach is to learn transparent \emph{surrogates} from the original models~\cite{adadi2018}.
A major challenge in this approach is balancing the fidelity of the surrogate to the original with the interpretability of the surrogate. 
Surrogate models that are generated stochastically can additionally struggle to provide consistent representations for the same policy. 

\begin{figure}[t]
    \centering
    \includegraphics[width=0.99\columnwidth]{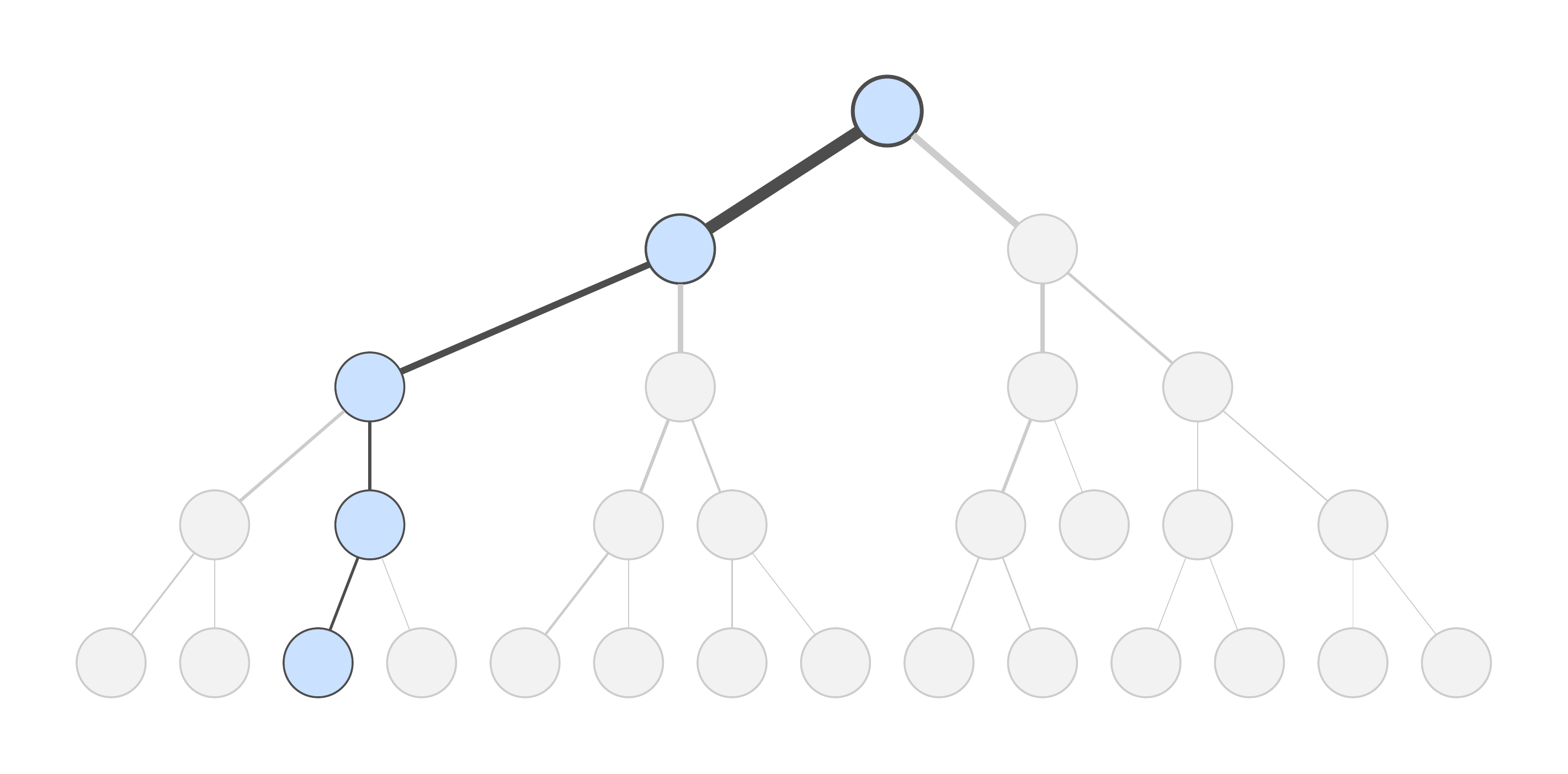}
    \caption{Surrogate Policy Tree. The figure shows a surrogate policy tree for the vaccine planning task, a finite-horizon MDP with 8 discrete actions. Each node represents an action and the states that led to it. Node borders and edge widths are proportional to the probability of encountering that node or edge during policy execution. The most likely trajectory from root to leaf is shown in blue.}
    \label{fig: graph unlabeled}
\end{figure}
% \begin{figure}[t]
%     \centering
%     \includegraphics[width=0.95\columnwidth]{figs/tree_mapping.pdf}
%     \caption{Policy Tree Mapping: This figure illustrates how the tree model approximates the baseline policy $\pi$ in a subset of the state space reachable from an initial point $s_0$. The orange represents the state space of the task $\mathcal{S}$ and the blue represents the reachable space $\mathcal{S}_R$ under $\pi$. The solid outline figure shows the space considered by the search tree.}
%     \label{fig:tree mapping}
% \end{figure}
% Neural networks are not globally transparent due to their high number of parameters and non-linear interactions. 
% Some methods have been proposed to constrain the architecture and training of neural networks such that the resulting policies extract human interpretable features and are intrinsically transparent.
% The training and architecture constraints, however, can lead to performance degradation compared to an unconstrained policy. 
% More commonly, methods learn transparent \emph{surrogates} from the original~\cite{adadi2018} models.
% A major challenge in this approach is balancing the fidelity of the surrogate to the original with the transparency of the surrogate. 
% Local surrogate models that are generated with stochastic algorithms can additionally struggle to provide consistent representations from the same input set. 

In this work, we propose a method to develop transparent surrogate models as local policy trees. 
The resulting trees encode an intuitive plan of future actions with high-fidelity to the original policy. 
% Tree build algorithm parameters can be tuned to constrain the tree size and expected fidelity loss. 
The proposed approach allows users to specify tree size constraints and fidelity targets.
The method is model-agnostic, meaning that it does not require the original policy to take any specific form.
Though this work was motivated by neural networks, the proposed approach may be used 
% to encode trees for 
with any baseline policy form. 

% The tree surrogates can be used as policies during task execution in which the course of action followed by an agent is guaranteed to be along one of the finite branches of the tree.  
During execution of a tree policy, the actions taken by an agent are guaranteed to be along one of the unique paths from the root. 
Experiments on a simple grid world task highlight the impact of algorithm parameters on tree behavior. 
The experiments also show that using the trees as a receding-horizon policy maintains good performance relative to the baseline model.
Additional experiments on more complex infrastructure planning and cyber-physical security domains demonstrate the potential utility of this approach in real-world tasks. 
\section{Background}
Despite its importance in machine learning, there is no precise qualitative or mathematical definition of interpretability. 
In the literature, interpretability is commonly used to describe a model with one of two characteristics. 
The first is explainability, defined by~\citet{miller2019} as the degree to which a human can understand the cause of a decision.
The second characteristic is predictability, which is defined by~\citet{kim2016} as the degree to which a human can consistently predict a model’s result.
In this work, we will refer to models that are either explainable or predictable as interpretable. 
% a model that possesses either of these traits as interpretable or transparent. 

% We may also use the term transparency in place of interpretability.
% An interpretable model is distinct from an explainable decision, which can be defined as the degree to which a human can understand the cause of a decision~\cite{miller2019}. 
Techniques for model interpretability can vary greatly. 
We characterize the methods presented in this work using the taxonomy proposed by~\citet{adadi2018} as model-agnostic, local surrogate methods.
Model-agnostic methods are those that can be may be applied to any model with the appropriate mapping from input to output. 
Surrogate modeling techniques distill the behavior of a complex \emph{baseline} model into a more transparent surrogate. 
% We refer to the model from which a surrogate is developed as the \emph{baseline}. 
Local methods provide interpretability that is valid only in the neighborhood of a target input point. 
As a result, they tend to maintain higher fidelity to the original model in the acceptable region than models that attempt to provide global interpretations.

\subsection{Markov Decision Processes}
The control tasks in this work are assumed to satisfy the Markov assumption and may be modeled as either Markov decision processes (MDPs) or partially observable Markov decision processes (POMDPs). 
MDPs are defined by tuples $(\mathcal{S}, \mathcal{A}, T, r, \gamma)$, where $\mathcal{S}$ and $\mathcal{A}$ are the state and action spaces, respectively, $T(s' \mid s, a)$ is the transition model, $r(s, a, s')$ is the reward function, and $\gamma$ is a time discount factor. 
POMDPs are modeled by the same tuple with the addition of the observation space $\mathcal{O}$ and observation model $Z(o \mid s, a)$. 
A policy is a function that maps a state to an action in MDPs or a history of observations to an action in POMDPs. 
To solve a MDP is to learn the policy $\pi: \mathcal{S} \to \mathcal{A}$ that maximizes the expected sum of discounted rewards 
\begin{equation}
    V(s_t) = \mathbb{E}\Bigg[\sum_{\tau=t} \gamma^{\tau-t} r\big(s_\tau, \pi(s_\tau)\big)\Bigg]
\end{equation}
over all states. 
Learning an optimal policy is equivalent to learning the optimal action value function 
\begin{equation}
    Q(s_t, a_t) = \mathbb{E}\Bigg[r(s_t, a_t) + \sum_{\tau=t+1} \gamma^{\tau-t} r\big(s_\tau, \pi^*(s_\tau)\big)\Bigg]
\end{equation}
where $\pi^*$ is the optimal policy. 
% It is common in reinforcement learning to learn to approximate the expected action value function $Q(s,a)$ instead of learning a policy directly, as in the case of DQN~\cite{mnih2015}. 
When learning an action value function estimator, the effective policy is
\begin{equation}
    \pi(s) \gets \arg\max_{a \in \mathcal{A}} \hat{Q}(s, a)
\end{equation}
where $\hat{Q}(s, a)$ is the learned approximator.
These problems can be solved using a variety of methods such as dynamic programming, Monte Carlo planning, and reinforcement learning~\cite{kochenderfer2022}. % TODO check cite
Similarly, various function types can be used to model the policy or value function estimator. 
Neural networks are commonly used as policies for complex tasks.

\section{Related Work}
There has been a wealth of prior work in the field of explainable and interpretable artificial intelligence. 
The book by~\citet{molnar2019} provides an overview of model interpretability techniques for general machine learning methods. 
Methods for specific models and tasks have also been proposed. 
\citet{puiutta2020} provides a survey of recent work in explainable reinforcement learning. 
The discussion in this section is restricted to methods relevant to deep reinforcement learning.
Methods requiring domain specific representations~\cite{verma2018}
% or non-neural policy representations 
are not considered.

A common technique for model-agnostic surrogate modeling is to learn a surrogate model of an inherently interpretable class. 
An useful example is locally interpretable model-agnostic explanations (LIME)~\cite{ribeiro2016}.
LIME learns sparse linear representations of a target policy at a specific input by training on a data set of points near the target point.
While the resulting linear functions are interpretable, they are often not consistent and small variations in the training data can result in drastically different linear functions.

Several methods have been proposed to distill neural network policies to decision tree surrogates.
Linear model U-trees~\cite{liu2018} and soft decision trees~\cite{coppens2019} take similar approaches to tree representation and learning.
Decision trees are global policy representations that map input observations to output actions by traversing a binary tree. 
Actions can be partially understood by inspecting the values at each internal node. 
Unfortunately, the understanding provided by decision trees can be limited because the mappings still pass the input observation through several layers of affine transforms and non-linear functions. 
Further, both methods empirically showed significant performance loss compared to the baseline policies.

Structural causal models~\cite{madumal2020} also try to learn an inherently interpretable surrogate as a directed acyclic graph (DAG). 
The learned DAG represents relations between objects in the task environment that can be easily understood by humans. 
The DAG structure, however, must be provided for each task, and the fidelity is not assured. 

Our work proposes tree representations that provide intuitive maps over future courses of action. 
When used as a policy, the realized course of action is guaranteed to be along the branches of the tree. 
In this way, the method is similar to methods of neural network verification that seek to provide guarantees of network outputs over a given set of inputs.
\citet{liu2021} provides an overview of verification for general neural networks. 
Additional works have been proposed specifically for verification of neural network control policies~\cite{sidrane2021}.

The methods proposed in this work also resemble Monte Carlo tree search (MCTS) algorithms~\cite{kocsis2006}. 
MCTS methods search for an optimal action from a given initial state or belief by sampling trajectories using a search policy. 
% MCTS can be used to solve fully observable Markov decision problems~\cite{kocsis2006} and partially observable Markov decision problems~\cite{Silver2010, sunberg2018}.
The result of an MCTS search is a tree of trajectories reachable from the initial state. 
The tree trajectories, however, are not limited to those reached under a fixed policy, but rather by a non-stationary tree search policy. 
The resulting tree cannot be used to interpret or predict future policy behavior. 
Trees from planners using UCB exploration~\cite{auer2002} tend to grow exponentially with search depth $h$ as $O(|\mathcal{S}|^h|\mathcal{A}|^h)$. %, where $\mathcal{S}$ and $\mathcal{A}$ are the state and action spaces, respectively. 
\section{Proposed Method}
% Local surrogate, Interpretable Predictability, control fidelity vs tree size, model agnostic, forward operation with guarantees, finite tree size, MC performance bounds
We present model-agnostic methods to represent policies for both MDPs and POMDPs as interpretable tree policies.
Trees are inherently more interpretable than high dimensional models like neural networks. 
The proposed method uses the baseline policy to generate simulations of trajectories reachable from a given initial state. 
The trajectories are then clustered to develop a width-constrained tree that represents the original policy with good fidelity. 
The methods assume that baseline policies return distributions over actions or estimates of action values for a given state. 
Building trees also requires a generative model of the task environment.

In addition to representing the future policy, the tree also provides useful statistics on expected performance such as likelihood of following each represented trajectory. 
The tree may be used as a policy during task execution with guarantees on expected behavior. 
We present methods to generalize to states not seen during tree construction by using the tree as a constraint on the baseline policy. 
% We show that trees may be incorporated into the control loop during task execution as receding-horizon controllers with guaranteed behavior. 

\subsection{Local Tree Policy}
\begin{figure*}[ht!]
    \centering
    \begin{subfigure}[t]{0.56\textwidth}
        \centering
        \includegraphics[width=\textwidth]{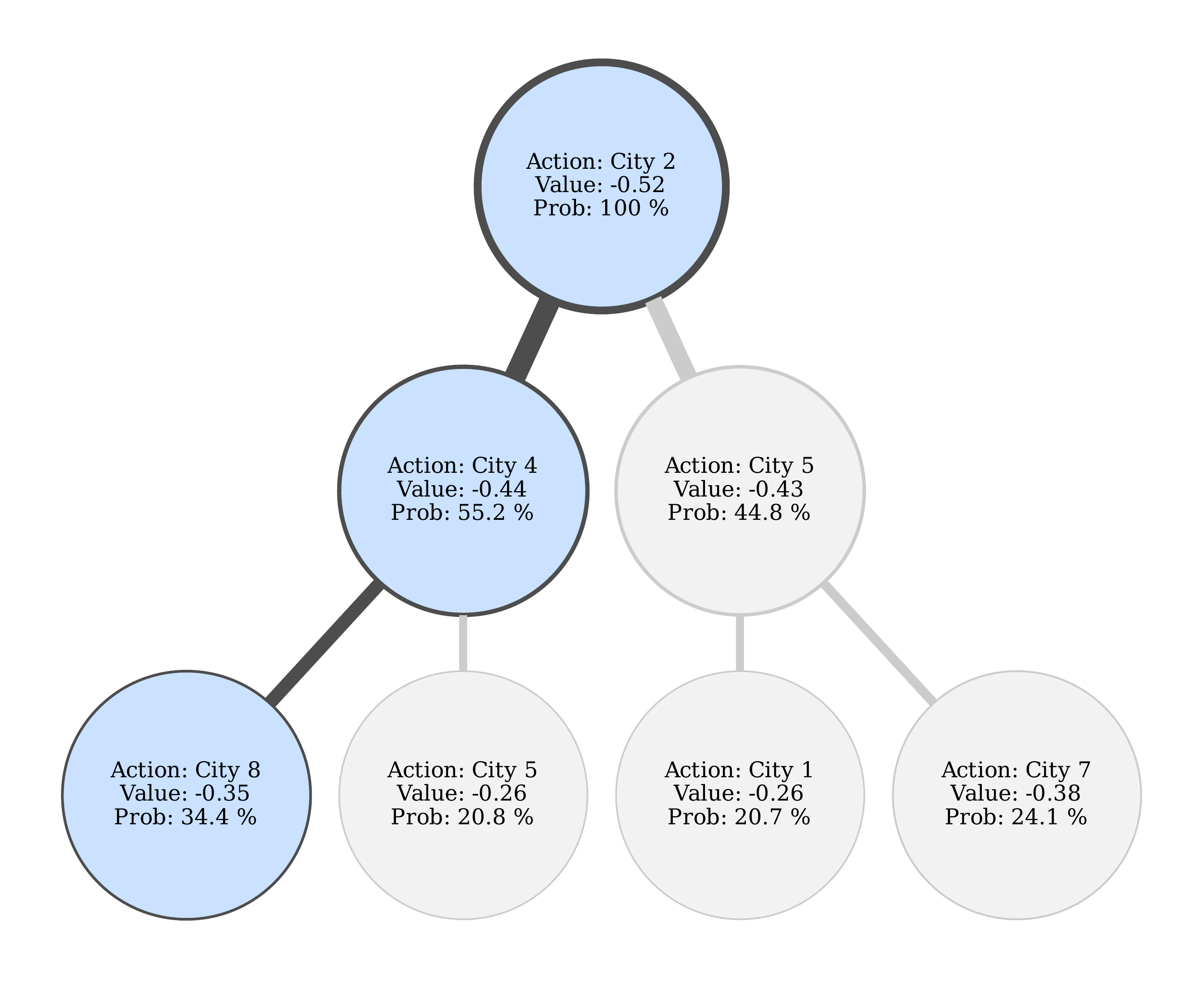}
        \caption{}\label{fig: detail a}
    \end{subfigure}
        \begin{subfigure}[t]{0.43\textwidth}
        \centering
        \includegraphics[width=\textwidth]{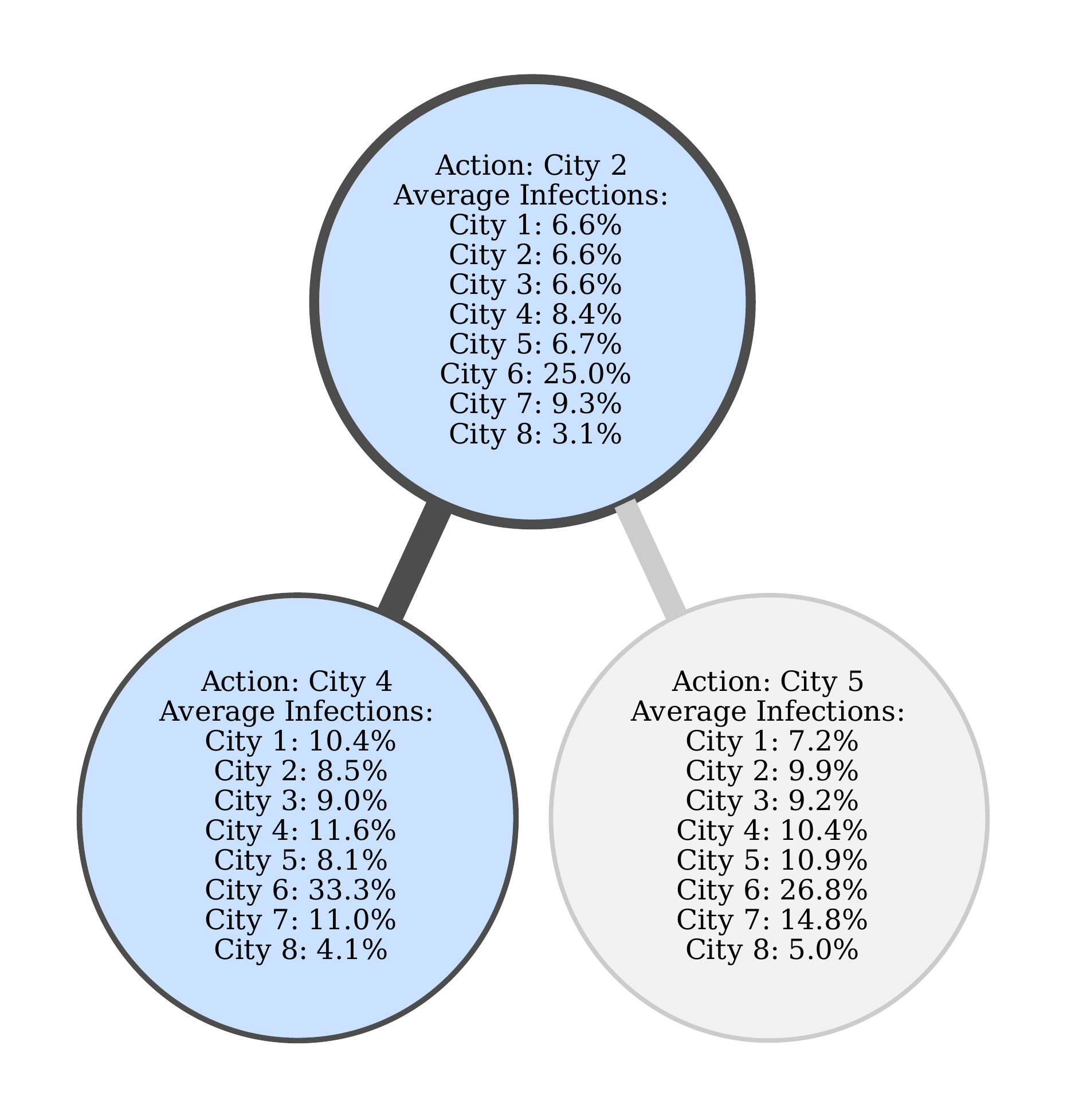}
    \caption{}\label{fig: detail b}
    \end{subfigure}
    \caption{Detailed Tree Views. (a) The figure shows the first three layers of the vaccine surrogate policy tree. Each node provides its action, estimated value, and probability of reaching it. (b) The figure shows the first two tree layers with the mean of the observations leading to each action node displayed.
        % As a result, the cumulative trajectory probability at a given level of the tree is not necessarily equal to 1.
        }
    \label{fig: example graph}
\end{figure*}
% \begin{figure}[t]
%     \centering
%     \includegraphics[width=0.99 \columnwidth]{figs/vaccine_graph.pdf}
%     \caption{Example Policy Tree. The figure shows a visualization of a surrogate policy tree for the vaccine planning task, a finite-horizon MDP with 8 discrete actions. Each node shows the action taken, the expected value, and the probability of reaching that node. Node borders and edge widths are proportional to the probability of encountering that node or edge during policy execution. The most likely trajectory from root to leaf is shown in blue. To prevent over-fitting, the tree is not expanded on nodes with less than 15 \% probability of being reached. 
%     % As a result, the cumulative trajectory probability at a given level of the tree is not necessarily equal to 1.
%     }
%     \label{fig: example graph}
% \end{figure}
% \begin{figure}[h]
%     \centering
%     \includegraphics[width=0.99 \columnwidth]{figs/vaccine_detail_graph.pdf}
%     \caption{
%     }
%     \label{fig: example graph}
% \end{figure}
Before describing the tree construction algorithm, we first present the local tree policies that it produces. 
A policy tree is a rooted polytree where nodes represent actions taken during policy execution. 
An example tree policy for a stochastic, fully observable task is shown in~\cref{fig: graph unlabeled}.
Each node of the tree represents an action $a_t$ taken at time $t$.
% A node's parent gives the action that necessarily preceded it.
The root action of each tree is the action recommended by the policy at the initial state.
Each path from the tree root to a leaf node gives a trajectory of actions $a_t, a_{t+1}, \dots, a_{t + h}$ that the agent may take during policy execution up to some depth $h$.
Trajectories leading to terminal conditions before $h$ steps result in shallower tree branches. 

Policy trees are interpretable representations of the future behavior of a policy that give explicit, quantitative predictions of future trajectories.
Each node provides an estimate of the probability that the action sequence up to that node will be taken $P(a_0, \dots, a_t \mid s_0, \pi)$ and estimates of the policy value $Q(s, a)$.
% These estimates can also be used to calculate additional values, such as the conditional probability of taking an individual action $P(a_t \mid a_0, \dots, a_{t-1}, s_0, \pi)$.
% Nodes also provide estimates of the policy value $Q(s, a)$ for each action and examples of the states or observations that would cause that action to be taken.
Each node also contains a set of example states or observations that would result in that action. 

\Cref{fig: example graph} provides more detailed views of the policy tree in~\cref{fig: graph unlabeled}.
The trajectory probabilities and value estimates shown in~\cref{fig: detail a} are calculated during tree construction and may be presented for any surrogate policy.
The states in the example problem $s_t \in \mathbb{R}^{n}$ are real vectors. 
Each node in~\cref{fig: detail b} shows the mean of that node's state values.
While the mean state value is useful for understanding this problem, it may not be useful in all problems.
For example, the mean value may be meaningless for problems with discrete state spaces.
Methods to compactly represent a node's set of states or observations cannot be generally defined and instead should be specified for each problem. 
% Each node in~\cref{fig: detail b} shows a subset of the mean state value giving the average fraction of infected population in each city.
% General methods to compactly represent each node's set of state or observations are not provided because the form of the observation or state can vary between problems.

% Policy trees may also be used as constraints on the baseline policy during execution. 
% This is described in further detail in later sections. 
% \begin{figure}[h]
%     \centering
%     \includegraphics[width=0.9\columnwidth]{figs/reachable_mapping.pdf}
%     \caption{Local Surrogate Fidelity. This figure illustrates the representation neighborhood of the policy tree representation compared to a baseline perturbation-based surrogate model. The orange represents the state space of the task and the blue represents the reachable space from the given initial state to the end of the episode under baseline policy $\pi$. The solid outline figure shows the space considered by the search tree, while the dashed circle shows an area modeled by a LIME approach. While the areas enclosed by the two figures are equivalent, much more of the reachable space is considered by the tree approach.}
%     \label{fig: reachable map}
% \end{figure}
Trees are an intuitive choice for the local surrogate model of a control policy. 
Local surrogate methods seek to provide simple approximate models that are faithful to the original policy in a space around a target point. 
For control tasks, often only predictions of \emph{future} behavior are useful. 
Trees provide compact surrogate models by only representing behavior in states that are forward-reachable from the current state. 
This is in contrast to methods that define local neighborhoods by arbitrarily perturbing the initial point~\cite{ribeiro2016}.
In these cases, explanatory capacity is wasted on unnecessary states. 
% In these cases, the surrogate must have increased complexity to model behavior in unreachable spaces or have diminished fidelity.

\subsection{Tree Build Algorithm}
\begin{figure*}[t]
    \centering
    \begin{subfigure}[b]{0.32\textwidth}
        \centering
        \includegraphics[width=\textwidth]{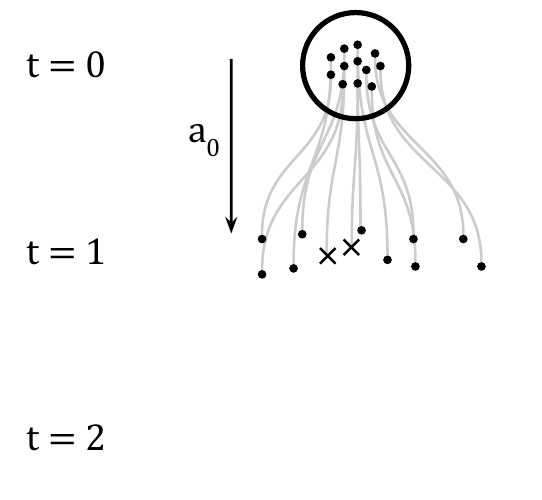}
        \caption{}\label{fig: build 1}
    \end{subfigure}
    \begin{subfigure}[b]{0.32\textwidth}
        \centering
        \includegraphics[width=\textwidth]{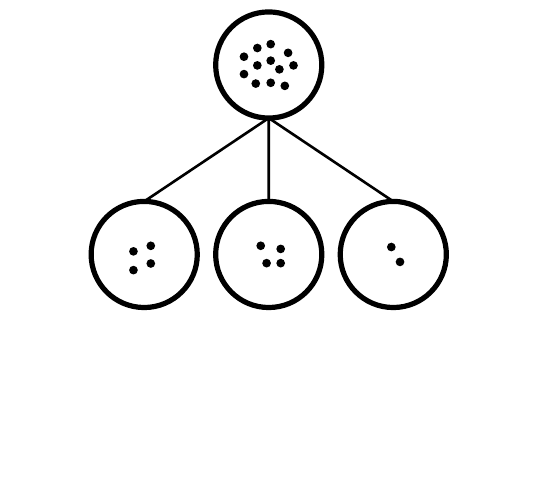}
        \caption{}\label{fig: build 2}
    \end{subfigure}
    \begin{subfigure}[b]{0.32\textwidth}
        \centering
        \includegraphics[width=\textwidth]{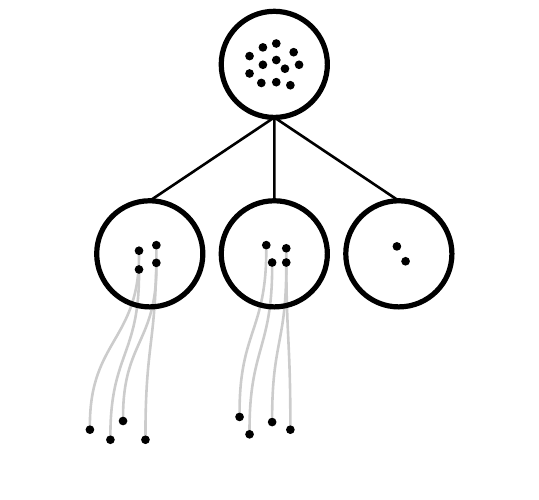}
        \caption{}\label{fig: build 3}
    \end{subfigure}
    \caption{Tree build example. (a) An initial set of particles is sampled from the given initial state or belief. These particles are advanced in the simulation using the action at the root node $a_0$. Particles reaching terminals states are marked with $\times$. (b) Non-terminal particles are clustered into new action nodes. (c) Action nodes with a sufficient number of particles are advanced another time step. This process will continue for each action node until all particles terminate or reach a maximum depth.}\label{fig: tree build}
\end{figure*}
Trees are built by simulating multiple executions of the baseline policy from the given initial state or belief and clustering them into action nodes.
The process is illustrated in~\cref{fig: tree build}.
Each simulation is represented by a collection of particles $p_t$ for each time $t$ along the simulation trajectory.
Particles are tuples $(s_t, r_t)$ of the state at time $t$ and reward received from $t-1$ to $t$. 
For POMDPs, particles also record the observation $o_t$ and belief $b_t$. 

Tree construction begins by first generating a set of particles for the initial state or belief.  
For the initial step, all particles are assigned to the root action node $an_0$, which takes the action given by the baseline policy $a_0 = \pi(s_0)$ for MDPs or $a_0 = \pi(b_0)$ for POMDPs.
The particles are then advanced through one simulation time step to produce the particles for $t+1$, as shown in~\cref{fig: build 1}.
The $t+1$ particles that did not enter terminal states are clustered to new action nodes as shown in~\cref{fig: build 2}.
This process continues until a terminal state is encountered or a fixed depth limit is met. 
Action nodes that do not have at least $n_{\min}$ particles are not expanded further to limit over-fitting, where $n_{\min}$ is specified by the user. 
% % Particles representing sampled environment instances are simulated from the root node using the action $a_0$
% % The tree encodes a deterministic surrogate of a baseline policy. % deterministic policy $\pi:\mathcal{S} \mapsto \mathcal{A}$. 
% % Trees can be constructed for policies over MDPs or POMDPs, 
% We present algorithms to construct tree surrogates for policies over MDPs and POMDPs.
% The majority of functions are shared between both algorithms. 
% Functions exclusive to MDPs and POMDPs are shown in~\cref{BuildMDP} and~\cref{BuildPOMDP}, respectively.  
% % A slightly different approach is taken to sample trajectories at the root for MDPs and POMDPs.
% Building trees for determinsitic policies over deterministic MDPs will result in trees with a single branch.  
% Surrogate trees can also be built for POMDP policies using a slightly modified approach which is shown in the appendix. %

% The points of entry for tree construction are the \textsc{Build} functions, presented in~\cref{BuildMDP} and~\cref{BuildPOMDP} for MDPs and POMDPs, respectively.
The point of entry for tree construction is the \textsc{Build} function, presented in~\cref{BuildMDP} for MDPs.
The initial particle set is created and clustered into the root action node.
Each root particle for an MDP policy is initialized with the same state $s_0$. 
Each action node is a tuple $(P, a, Ch)$, where $P$ is the node's set of particles, $a$ is the action taken from that node, and $Ch$ is the set of child nodes. 
The \textsc{Build} function for POMDPs follows the same procedure as for MDPs, though states and observations for the initial particle set are sampled from the initial belief $b_0$.
\begin{algorithm}[htb]
\caption{Build MDP}\label{BuildMDP} 
\begin{algorithmic}[1]
    \Procedure{BuildMDP}{$s_0$, $\pi$, $n$, $d_{max}$}
        \State $P \gets \emptyset$
        \State $Ch \gets \emptyset$
        \State $d \gets 0$
        \For {$i \in 1 : n$}
            \State $P \gets P \cup \{ (s_0, 0) \}$
        \EndFor
        % \State $a_0 \gets \arg\max_{b} \pi(s_0)$
        % \State $q_0 \gets \max_{b} \pi(s_0)$
        \State $p_0 \gets \pi(s_0)$
        \State $a_0 \gets \arg\max_{a \in \mathcal{A}} \pi(s_0)$
        \State $\text{root} \gets \text{Node}(P, a_0, p_0, Ch)$
        \State $\text{root} \gets \textsc{Rollout}(\text{root}, \pi, d)$
        \State \Return $\text{root}$
    \EndProcedure
\end{algorithmic}
\end{algorithm}

% \begin{algorithm}[htb]
% \caption{Build POMDP}\label{BuildPOMDP} 
% \begin{algorithmic}[1]
%     \Procedure{BuildPOMDP}{$s_0$, $o_0$, $b_0$, $\pi$, $n$, $d_{max}$}
%         \State $P \gets \emptyset$
%         \State $Ch \gets \emptyset$
%         \State $d \gets 0$
%         \For {$i \in 1 : n$}
%             \State $s_0 \sim b_0$
%             \State $P \gets P \cup \{ (s_0, 0, o_0, b_0) \}$
%         \EndFor
%         % \State $a_0 \gets \arg\max_{b} \pi(s_0)$
%         % \State $q_0 \gets \max_{b} \pi(s_0)$
%         % \State $p_0 \gets \pi(s_0)$
%         \State $a_0 \gets \arg\max_{a \in \mathcal{A}} \pi(s_0)$
%         \State $\text{root} \gets \text{Node}(P, a_0, Ch)$
%         \State $\text{root} \gets \textsc{Rollout}(\text{root}, \pi, d)$
%         \State \Return $\text{root}$
%     \EndProcedure
% \end{algorithmic}
% \end{algorithm}

Particles are advanced through recursive calls to the \textsc{Rollout} function~(\cref{Rollout}). 
Each time \textsc{Rollout} is called from an action node $an$, it proceeds if the size of the node particle set $an_P$ exceeds the minimum threshold and the depth limit $d_{\max}$ has not been exceeded.
If these conditions are met, each particle is advanced by calling the simulator $\textsc{Gen}(s, a)$ using that particle's state and the node action. 
The set of new particles are then grouped into new action nodes by the \textsc{Cluster} function.
\textsc{Rollout} is recursively called on each new action node and the returned node is added to the child node set $an_{Ch}$.
\begin{algorithm}[ht]
\caption{Rollout}\label{Rollout} 
\begin{algorithmic}[1]
    \Procedure{Rollout}{$an$, $\pi$, $d$}
        \If{$|an_{P}| \geq n_{min}$ and $d < d_{max}$}
            \State $P \gets \emptyset$
            % \State $a \gets an_a$
            \For {$p \in an_P$}
                \State $s \gets p_s$
                \State $s', o', r', \text{done} \gets \textsc{Gen}(s, an_a)$ 
                \If {not done}
                    % \State $a' \gets \arg\max_{b} \pi(s')$
                    % \State $q' \gets \max_{b} \pi(s')$
                    % \State $\text{node}' \gets \text{Node}(\{(s', r')\}, a', q', \{ \ \})$
                    \State $p' \gets \textsc{Particle}(s', o', r')$
                    \State $P \gets P \cup \{ p' \}$
                \EndIf
            \EndFor
            \State $C \gets \textsc{Cluster}(P, \pi)$
            % \State $C' \gets \{ \ \}$
            \For {$c \in C$}
                \State $c' \gets \textsc{Rollout}(c, \pi, d + 1)$
                \State $an_{ch} \gets an_{ch} \cup \{ c' \}$
            \EndFor
            % \State $\text{node} \gets \textsc{UpdateChildren}(\text{node}, C')$
        \EndIf
        \State \Return \text{node}
    \EndProcedure
\end{algorithmic}
\end{algorithm}

The action nodes of the tree $T$ encode a deterministic policy for the sampled particles, $T: \mathcal{S}_R \to \mathcal{A}$, where $\mathcal{S}_R$ is the set of states contained in the particle set. 
The particles are clustered into action nodes to minimize how much this policy deviates from the baseline policy $\pi$ while meeting constraints on tree size. 
To achieve this, we developed a clustering algorithm that approximately solves for the optimal clustering through recursive greedy optimization.

The recursive clustering approach is presented in~\cref{Cluster}.
The algorithm progressively clusters particles into an increasing number of nodes $k$ until a distance measure $\delta$ between the actions assigned by the tree and the baseline policy is less than a threshold $\delta^*$ or the maximum number of nodes $c_{\max}$ is exceeded. 
\begin{algorithm}[ht]
\caption{Recursive Cluster}\label{Cluster}
\begin{algorithmic}[1]
    \Procedure{Cluster}{$P$, $\pi$}
        \State $k \gets 1$
        \Repeat
            \State $C, \delta \gets \textsc{GreedyCluster}(P, k, \pi)$
            \State $k \gets k + 1$
        \Until{$\delta \leq \delta^*$ or $|C| \geq c_{\max}$}
        \State \Return $C$
    \EndProcedure
\end{algorithmic}
\end{algorithm}

Clusters are assigned according to a greedy heuristic process shown in~\cref{GreedyCluster}.
In this approach, the complete set of actions assigned to each particle under the baseline policy $A^U$ is ranked according to frequency by the \textsc{UniqueActions} function. 
Action nodes for each of the top $k$ actions and all particles assigned that action by the baseline policy are clustered. 
Any particles not clustered to a node at the end of this process are assigned to the previously formed node that minimizes the distance between that action and the action assigned by the baseline policy. 
\begin{algorithm}[ht]
\caption{Greedy Cluster}\label{GreedyCluster}
\begin{algorithmic}[1]
    \Procedure{GreedyCluster}{$P$, $k$, $\pi$}
        \State $C \gets \emptyset$
        \State $\delta \gets 0$
        \State $A^U \gets \textsc{UniqueActions}(P, \pi)$
        \For{$i \in 1 : k$}
            \State $a \gets A^U[i]$
            \State $P^A \gets \{ p \in P \mid \arg\max_{a \in \mathcal{A}}\pi(p_s) = a \}$
            \State $C \gets \{ \textsc{Node}(P^A, a, \emptyset) \} \cup C$
            \State $P \gets P \setminus P^A$
        \EndFor
        \For{$p \in P$}
            \State $an \gets \arg\min_{an' \in C} \textsc{Distance}(p, an', \pi)$
            \State $\delta \gets \delta + \textsc{Distance}(p, an, \pi)$
            \State $an_P \gets \{ p \} \cup an_P$
        \EndFor
        \State \Return $C$, $\delta$
    \EndProcedure
\end{algorithmic}
\end{algorithm}

The distance metric may be any appropriate measure assigned by the user for the given policy type. 
For action value function policies, an effective distance measure for an action $a$ would be 
\begin{equation}\label{eq: delta dqn}
    \delta = \| \hat{Q}(s, a) - \max_{a' \in \mathcal{A}} \hat{Q}(s, a') \| 
\end{equation}
which intuitively gives the predicted sub-optimality of the action.
For stochastic policies which output distributions over actions, the difference in action probability would be an appropriate distance.

\subsection{Tree Policy Control}\label{sec: tree execution}
To use the tree as a policy during task execution, it must be able to generalize to states not encountered in the particle set of the tree. 
To do this, we propose a simple method that uses the baseline policy, constrained by the tree at each time step. 
For a tree constructed for a state $s_0$, the policy will always take the action at the root node $an_0$.
For all remaining steps, the policy will return
\begin{equation}
    a \gets \arg\max_{a' \in \mathcal{A}^T} \pi(s_t)
\end{equation}
where $\mathcal{A}^T$ is the set of actions in the child set of the preceding action node $an_{t-1}$.
Intuitively, the agent will take the best action predicted by the baseline policy that is included in the tree. 
Building a new tree each time a leaf state is encountered allows the tree policies to be run in-the-loop for long or infinite-horizon problems.

\section{Experiments}
We ran experiments to test the performance of the proposed approach. 
One set of experiments are conducted on a simple grid world task. 
These experiments were designed to quantitatively measure the effect of various algorithm parameters and environment features on tree size and fidelity. 
Two additional experiments were run on more complex tasks that demonstrate the utility of the approach on real-world motivating examples. 
The first of these is multi-city vaccine deployment planning. 
In large-scale infrastructure planning tasks such as this, it is often necessary to have a reasonable prediction of all steps of the plan in order to gain stakeholder trust.
The second task is a cyber security agent that provides recommendations to a human analyst to secure a network against attack. 
Interpretable policies are important for tasks with human oversight and cooperation.  

We implemented the proposed algorithm with the distance function defined in~\cref{eq: delta dqn} in Python. % and trained baseline policies using various methods. 
% We implemented the distance function for value function estimation policies defined in~\cref{eq: delta dqn}. 
Neural network training was done in PyTorch~\cite{paszke2015}.
% For each task, we trained a neural network using double DQN~\cite{hasselt2010} with $n$-step returns in PyTorch~\cite{paszke2015}.
Source code, full experiment descriptions, and results are available in the Appendix. 
% Source code for the algorithm implementation and all test environments is also provided. 
% \begin{table}[ht] % Maybe change to best performance
%     \centering
%     \begin{tabular}{l l r r}
%         \toprule
%         Task & Solver & Rel Change (\%) & Leaf Depth \\
%         \midrule
%         Grid World & VI & $-4.4 \pm 1.0$ & $4.2 \pm 0.0$ \\
%         % Grid World & SARSA & $X \pm Y$ & $X \pm Y$ \\
%         Grid World & DQN & $-8.3 \pm 1.2$ & $5.0 \pm 0.0$ \\
%         \midrule
%         Vaccine Rollout & DQN & $X \pm Y$ & $X \pm Y$ \\
%         Cyber Security & DQN & $X \pm Y$ & $X \pm Y$ \\
%         \bottomrule
%     \end{tabular}
%     \caption{Results Summary}
%     \label{tab: results summary}
% \end{table}
\subsection{Grid World}
In the grid world task, an agent must navigate a discrete, 2D world to reach a goal state while avoiding trap states.
% The grid is $20 \times 20$ and has 10 goal states and 10 trap states.
The agent may take one of $n$ total actions to move between $1$ and $\lfloor \frac{n}{4}\rfloor$ units in any of the four cardinal directions on the grid.
The agent moves in the intended direction with probability $p$ and takes a random action otherwise.
% The gridworld environment is a standard $20 \times 20$ gridworld with $10$ goal states and $10$ trap states. 
% Each time step, the agent receives a reward of $-1$ plus $10$ if it reaches a goals state or $-5$ to the reward while each trap state adds $-5$ to the reward.
To incentivize solving the problem quickly, the agent receives a cost of $-1$ at each time step. 
The agent gets a positive reward of $+10$ for reaching a goal state and a penalty of $-5$ for reaching a trap state.
The episode terminates when the agent reaches a goal state or when a maximum number of steps is reached. 
% The agent's actions are determine by the total number of actions, $n$. 
% The agent is able to move between $1$ and $\lfloor \frac{n}{4}\rfloor$ units in any of the $4$ cardinal directions on the grid. 
% Additionally, there is a transition probability, $p$ which represents the probability that the intended action is taken while there is a $1 - p$ probability of a random action. 
Because we know exact transition probabilities, we used discrete value iteration to learn a baseline policy~\cite{sutton1998}. 

We constructed surrogate trees from the baseline policy using various environment and tree-build algorithm settings.
For the environment, we swept over different values of the transition probability $p$ and the number of actions $n$.
For the tree build algorithm, we varied the total number of particles, the minimum particle count, the distance threshold $\delta^*$, and the maximum leaf node depth. 
Only one parameter was varied for each test, with all others held fixed. 
The baseline settings for the environment were $n=4$ actions and a successful transition probability of $p=0.9$.
The baseline tree build parameters are 1000 total particles, 250 minimum particles, distance threshold of 0.01, and maximum depth of 10.

To test each tree, we ran it as a policy from the initial state until it reached a leaf node.
The baseline policy was then used to complete the episode.
We tested 2,500 trees for each parameter configuration. 
Select results are shown~\cref{tab: grid results}.
The mean change in performance of the tree policy relative to the baseline is shown along with one standard error bounds. 
The average depth of leaf nodes is also shown, though standard error is omitted as each had $SE < 0.05$.
% The relative change of the tree policy  and the average leaf depth of the tree -- the standard deviation is 0.0 for all rows. 

From the transition probability sweep, we see that relative performance generally improves as transition probability increases
At $p=0.5$, both policies perform very poorly and the difference between the two is not significant as a result.
In the deterministic case, the surrogate tree perfectly represents the baseline policy. 
These results suggest that surrogate trees are better suited for tasks with low stochasticity. 
% We see that for the transition probability of $0.5$, both the baseline policy and the tree policy perform poorly as half of all moves are random. 
% Thus there is no significant difference between the two policies. 
% As the transition probability increases, the baseline policy performs better relative to the tree however when the transition probability is $1.0$, the value iteration solver achieves the maximum discounted reward as does the tree policy so there is no difference there.
\begin{table}[t]
    \centering
    \begin{tabular}{@{}l c c c@{}}
        \toprule
        Parameter & Value & Rel Change (\%) & Leaf Depth \\
        \midrule
        \multirow[c]{3}{*}{Trans. Prob.} & 0.5 & $4.5 \pm 4.7$ & $4.4$ \\ 
        & 0.7 & $-8.0 \pm 4.0$ & $4.4$ \\ 
        & \textbf{0.9} & $\mathbf{-4.4 \pm 1.0}$ & $\mathbf{4.2}$ \\ 
        & 1.0 & $0.0 \pm 0.0$ & $4.0$ \\ 
        \midrule
        \multirow[c]{3}{*}{No. Actions} & \textbf{4} & $\mathbf{-4.4 \pm 1.0}$ & $\mathbf{4.2}$ \\ 
        & 8 & $-2.6 \pm 0.4$ & $2.6$ \\ 
        & 16 & $-1.0 \pm 0.3$ & $1.5$ \\ 
        \midrule
        \multirow[c]{3}{*}{$\delta^*$} & 0.005 & $-5.7 \pm 1.0$ & $4.2$ \\ 
        & \textbf{0.01} & $\mathbf{-4.4 \pm 1.0}$ & $\mathbf{4.2}$ \\ 
        & 0.1 & $-3.1 \pm 1.0$ & $3.1$ \\ 
        \midrule
        \multirow[c]{3}{*}{Max Depth} & 3 & $-1.3 \pm 0.9$ & $2.0$ \\ 
        & 5 & $-2.9 \pm 1.0$ & $3.5$ \\ 
        & \textbf{10} & $\mathbf{-4.4 \pm 1.0}$ & $\mathbf{4.2}$ \\ 
        \bottomrule
    \end{tabular}
    \caption{Parameter Search Results. The change in task performance relative to the baseline policy performance and the average depth of leaf nodes in the trees are given. Values generated with the baseline settings are shown in bold.}
    \label{tab: grid results}
\end{table}

As we increase the number of actions, the difference in performance between the baseline and the tree policy decreases. 
The average leaf depth of the trees also decreases with more actions. 
This likely explains the improved performance, as shallower trees will transition back to the baseline policy sooner in the tests. 
% This could explain the trend as with a smaller leaf depth, the tree policy adopts the same actions as the baseline policy within a shorter number of time steps on average. 
% Thus as the number of actions grows, the leaf depth decreases and the tree policy performs better relative to the baseline policy. 
Another possible explanation is that as the number of actions grows, the cost of taking a sub-optimal action decreases. 
For example, with $4$ actions, if the optimal action is not taken, then the agent moves in a completely different direction than it should.
With $n>4$ actions, the agent may still move in the correct direction, though by more or less distance than optimal.
% However, with $16$ actions, if the optimal action is moving $3$ units east, then it is possible that a suboptimal action is taken by moving only $2$ units east and thus the difference in the discounted reward decreases. 
% For these environmental parameters, the key takeaways are that reducing stochasticity in the environment decreases the relative change between the baseline policy and the tree policy. Additionally, as we increase the number of actions, we gain finer control over the agent and in this case, we see that the relative performance of the tree policy improves.  

% The next two parameters are hyperparameters of the tree building algorithm itself. 
For the tree building algorithm parameters, we see that as $\delta^*$ increases, the depth of the leaf nodes also decreases. 
This is likely because higher values of $\delta^*$ leads to more aggressive node clustering and fewer branches. 
% nodes more frequently which reduces the number of branches. 
Similar to the trend observed by varying the number of actions, as the average leaf depth decreases, relative performance increases. 
Similarly, as the max depth increases, the tree is allowed to grow deeper and the relative performance drops. 
% However, we can also note that even when the maximum depth allowed is $10$, the average leaf depth is still $4.2$ as most particles terminate or are in nodes with fewer than than the minimum particle limit before we reach a tree depth of $10$. 
% Thus the main takeaway from the hyperparameter search is that both hyperparameters -- $\delta$ and max depth -- can influence the tree depth which is most directly tied to relative performance of the tree policy. 

\subsection{Vaccine Planning}
In the vaccine deployment task, an agent decides the order in which to start vaccine distributions in cities in the midst of a pandemic outbreak. 
Each step, the agent picks a city in which to start a vaccine program.
The spread of the disease is modeled as a stochastic SIRD model~\cite{bailey1975}, with portions of each city population being susceptible to infection (S), infected (I), recovered (R), or dead (D). 
% In cities with vaccine programs, some of the susceptible population skips infection and goes straight to recovery each time step. 
The $n$ cities are modeled as a fully connected, weighted graph, where weight $w_{ij}$ encodes the amount of traffic between city $i$ and $j$.
The state of city $i$ changes from time $t$ to $t+1$ as
\begin{align}
    dS^{(i)}_{t} &= -\beta \tilde{I}^{(i)}_t S^{(i)}_t - \alpha^{(i)}_t S^{(i)}_t +\epsilon^{S, (i)}_t \\
    dI^{(i)}_{t} &= \beta \tilde{I}^{(i)}_t S^{(i)}_t +\epsilon^{S, (i)}_t - \gamma I^{(i)}_t - \mu I^{(i)}_t \\
    dR^{(i)}_{t} &= \gamma I^{(i)}_t + \alpha^{(i)}_t S^{(i)}_t \\
    dD^{(i)}_{t} &= \mu I^{(i)}_t 
\end{align}
where $\beta$, $\gamma$, and $\mu$ are the mean infection, recovery, and death rates, respectively. 
The vaccination rate $\alpha^{(i)}_{t}$ of city $i$ at time $t$ and is equal to zero until an action is taken to deploy a vaccination program to that city.
The noise $\epsilon$ is sampled from a zero mean Gaussian. 
The effective infection exposure at city $i$ is defined as $\tilde{I}^{(i)} = \sum_j w_{ij}I_j$, where the sum is taken over all cities, and $w_{ii} = 1$.
Cities with closer index values will have higher weights, for example $w_{12} > w_{13}$.

Each episode is initialized with up to 10\% of each city infected and the remainder susceptible. 
One city is initialized with 25\% infected. 
The episode concludes after five cities have had vaccine programs started. 
The simulation is then run until all cities have zero susceptible or infected population.
The reward at each time step is $ r_t = a \sum_i dI^{(i)}_t + b \sum_i dD^{(i)}_t $, where $a$ and $b$ are parameters. 

We trained a neural network policy using double DQN~\cite{hasselt2010} with $n$-step returns.
% We trained a neural network policy with deep Q-learning to solve the task. 
% The network had three layers with 64 hidden units each. 
The trained policy achieved an average score with one standard error bounds of $-149.8 \pm 0.6$ over 100 trial episodes.
We built trees for 100 random initial states with 2000 particles and a minimum particle threshold of 250 and tested their performance as forward policies. 
The trees achieved an average score of $-152.2 \pm 0.8$ over the 100 trials, for an average performance drop of $1.6 \%$. 
To compare to a baseline surrogate modeling approach, we also trained and tested a LIME model~\cite{ribeiro2016} with 2000 samples.
The LIME average score was $-184.2 \pm 4.5$, for an average performance loss of $23.0 \%$.

The surrogate tree in~\cref{fig: example graph} provides an intuitive understanding of the learned policy. 
In~\cref{fig: detail b} we can see that the policy does not deploy vaccines to the most heavily infected cities first. 
It instead prioritizes cities with larger susceptible populations to give the vaccine time to take effect on a larger amount of the population. 
% The policy initially deploys to cities further from the heavily infected city. 
\subsection{Cyber Security}
\begin{figure}[t]
    \centering
    \includegraphics[width=0.99\columnwidth]{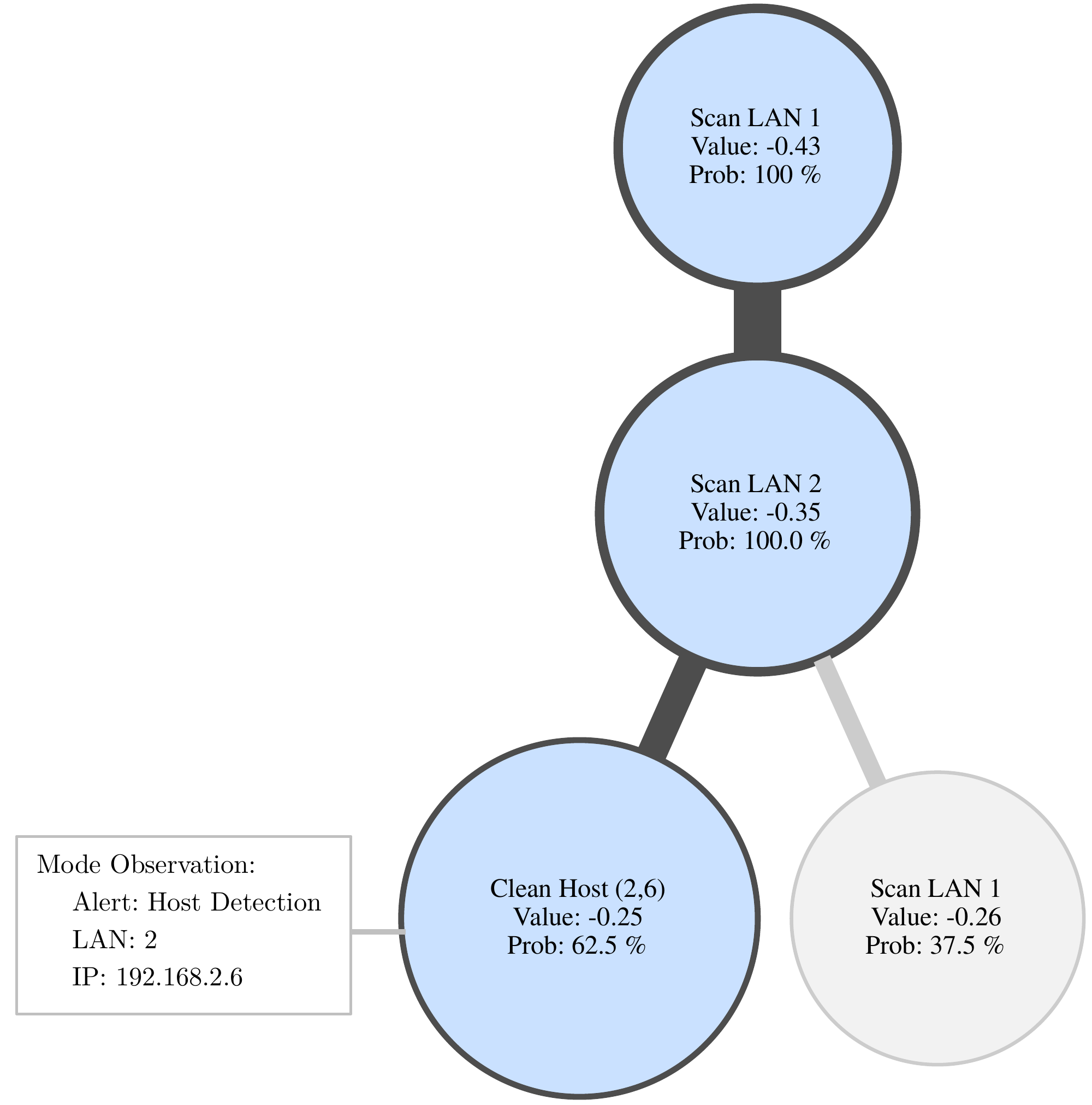}
    \caption{Cyber Security Policy Tree. The figure shows the first three layers of a surrogate tree for the cyber security task. The most frequent observation from the ``Clean Host" action node set is shown.}
    \label{fig: cyber tree}
\end{figure}
The cyber security task requires an agent to prevent unauthorized access to secure data server on a computer network. 
The computer network is comprised of four local area networks (LANs), each of which has a local application server and ten workstations, and a single secure data server. 
The compromise state of the network is not known may be observed through noisy alerts generated from malware scans. 
Workstations are networked to all others on their LAN and to the LAN application server. 
Servers are randomly connected in a complete graph. 
An attacker begins with a single workstation compromised and takes actions to compromise additional workstations and servers to reach the data server. 

The defender can scan all nodes on a LAN to locate compromised nodes with probability $p_{detect}$. 
Compromised nodes will also generate alerts without being scanned with low probability.
The defender can also scan and clean individual nodes to detect and remove compromise. 
The reward is zero unless the data server is compromised, in which case a large penalty is incurred.
The defender was trained using Rainbow DQN~\cite{hessel2018}.
% Local tree policies were constructed for the initial steps. 

Automated systems such as this are often implemented with a human in the loop. 
Policies that can be more easily interpreted are more likely to be trusted by a human operator. 
A surrogate tree for the neural network is shown in~\cref{fig: cyber tree}.
Unlike the baseline neural network, the policy encoded by this tree can be easily interpreted. 
The agent will continually scan LAN 1 in most cases, and will only clean a workstation after malware has been detected.
% Inspecting the states contained in these nodes reveals, for example, that a node is only cleaned if a prior scan revealed a sign of compromise.

\section{Conclusions}
In this work, we presented methods to construct local surrogate policy trees from arbitrary control policies. 
The trees are more interpretable than high-dimensional policies such as neural networks and provide quantitative estimates of future behavior. 
Our experiments show that, despite truncating the set of actions that may be taken at each future time step, the trees retain fidelity with their baseline policies.
Experiments demonstrate the effect of various environment and algorithm parameters on tree size and fidelity in a simple grid world.
Demonstrations show how surrogate trees may be used in more complex, real-world scenarios. 

The action node clustering presented in this work used a heuristic search method that provided good results, but without any optimality guarantees. 
Future work will look at improved approaches to clustering, for example by using a mixed integer program optimization.
We will also explore using the scenarios simulated to construct the tree to backup more accurate value estimates, and refine the resulting policy. 
Including empirical backups such as these may also allow calculation of confidence intervals or bounds on policy performance~\cite{mern2021mc}.

\bibliography{bib.bib}

\begin{thebibliography}{23}
\providecommand{\natexlab}[1]{#1}

\bibitem[{Adadi and Berrada(2018)}]{adadi2018}
Adadi, A.; and Berrada, M. 2018.
\newblock Peeking Inside the Black-Box: {A} Survey on Explainable Artificial
  Intelligence {(XAI)}.
\newblock \emph{{IEEE} Access}, 6: 52138--52160.

\bibitem[{Auer, Cesa{-}Bianchi, and Fischer(2002)}]{auer2002}
Auer, P.; Cesa{-}Bianchi, N.; and Fischer, P. 2002.
\newblock Finite-time Analysis of the Multiarmed Bandit Problem.
\newblock \emph{Journal of Machine Learning Research}, 47(2-3): 235--256.

\bibitem[{Bailey(1975)}]{bailey1975}
Bailey, N.~T. 1975.
\newblock \emph{The mathematical theory of infectious diseases and its
  applications}.
\newblock 2nd edition. Charles Griffin \& Company Ltd.

\bibitem[{Coppens et~al.(2019)Coppens, Efthymiadis, Lenaerts, Now{\'e}, Miller,
  Weber, and Magazzeni}]{coppens2019}
Coppens, Y.; Efthymiadis, K.; Lenaerts, T.; Now{\'e}, A.; Miller, T.; Weber,
  R.; and Magazzeni, D. 2019.
\newblock Distilling deep reinforcement learning policies in soft decision
  trees.
\newblock In \emph{International Joint Conference on Artificial Intelligence
  (IJCAI)}, 1--6.

\bibitem[{Hessel et~al.(2018)Hessel, Modayil, van Hasselt, Schaul, Ostrovski,
  Dabney, Horgan, Piot, Azar, and Silver}]{hessel2018}
Hessel, M.; Modayil, J.; van Hasselt, H.; Schaul, T.; Ostrovski, G.; Dabney,
  W.; Horgan, D.; Piot, B.; Azar, M.~G.; and Silver, D. 2018.
\newblock Rainbow: Combining Improvements in Deep Reinforcement Learning.
\newblock In \emph{AAAI Conference on Artificial Intelligence (AAAI)},
  3215--3222.

\bibitem[{Kim, Koyejo, and Khanna(2016)}]{kim2016}
Kim, B.; Koyejo, O.; and Khanna, R. 2016.
\newblock Examples are not enough, learn to criticize! Criticism for
  Interpretability.
\newblock In \emph{Advances in Neural Information Processing Systems
  (NeurIPS)}, 2280--2288.

\bibitem[{Kochenderfer, Wheeler, and Wray(2022)}]{kochenderfer2022}
Kochenderfer, M.~J.; Wheeler, T.~A.; and Wray, K.~H. 2022.
\newblock \emph{Algorithms for Decision Making}.
\newblock MIT Press.

\bibitem[{Kocsis and Szepesv{\'{a}}ri(2006)}]{kocsis2006}
Kocsis, L.; and Szepesv{\'{a}}ri, C. 2006.
\newblock Bandit Based {M}onte-{C}arlo Planning.
\newblock In \emph{European Conference on Machine Learning (ECML)}.

\bibitem[{Lipton(2018)}]{lipton2018}
Lipton, Z.~C. 2018.
\newblock The mythos of model interpretability.
\newblock \emph{Communications of the {ACM}}, 61(10): 36--43.

\bibitem[{Liu et~al.(2021)Liu, Arnon, Lazarus, Strong, Barrett, and
  Kochenderfer}]{liu2021}
Liu, C.; Arnon, T.; Lazarus, C.; Strong, C.~A.; Barrett, C.~W.; and
  Kochenderfer, M.~J. 2021.
\newblock Algorithms for Verifying Deep Neural Networks.
\newblock \emph{Foundations and Trends in Optimization}, 4(3-4): 244--404.

\bibitem[{Liu et~al.(2018)Liu, Schulte, Zhu, and Li}]{liu2018}
Liu, G.; Schulte, O.; Zhu, W.; and Li, Q. 2018.
\newblock Toward Interpretable Deep Reinforcement Learning with Linear Model
  U-Trees.
\newblock In \emph{European Conference on Machine Learning (ECML)}, volume
  11052, 414--429.

\bibitem[{Madumal et~al.(2020)Madumal, Miller, Sonenberg, and
  Vetere}]{madumal2020}
Madumal, P.; Miller, T.; Sonenberg, L.; and Vetere, F. 2020.
\newblock Explainable Reinforcement Learning through a Causal Lens.
\newblock In \emph{AAAI Conference on Artificial Intelligence (AAAI)},
  2493--2500.

\bibitem[{Mern and Kochenderfer(2021)}]{mern2021mc}
Mern, J.; and Kochenderfer, M.~J. 2021.
\newblock Measurable Monte Carlo Search Error Bounds.
\newblock \emph{Computing Research Repository}, abs/2106.04715.

\bibitem[{Miller(2019)}]{miller2019}
Miller, T. 2019.
\newblock Explanation in artificial intelligence: Insights from the social
  sciences.
\newblock \emph{Artificial Intelligence}, 267: 1--38.

\bibitem[{Mnih et~al.(2015)Mnih, Kavukcuoglu, Silver, Rusu, Veness, Bellemare,
  Graves, Riedmiller, Fidjeland, Ostrovski, Petersen, Beattie, Sadik,
  Antonoglou, King, Kumaran, Wierstra, Legg, and Hassabis}]{mnih2015}
Mnih, V.; Kavukcuoglu, K.; Silver, D.; Rusu, A.~A.; Veness, J.; Bellemare,
  M.~G.; Graves, A.; Riedmiller, M.~A.; Fidjeland, A.; Ostrovski, G.; Petersen,
  S.; Beattie, C.; Sadik, A.; Antonoglou, I.; King, H.; Kumaran, D.; Wierstra,
  D.; Legg, S.; and Hassabis, D. 2015.
\newblock Human-level control through deep reinforcement learning.
\newblock \emph{Nature}, 518(7540): 529--533.

\bibitem[{Molnar(2019)}]{molnar2019}
Molnar, C. 2019.
\newblock \emph{Interpretable Machine Learning}.
\newblock \url{https://christophm.github.io/interpretable-ml-book/}.

\bibitem[{Paszke et~al.(2019)Paszke, Gross, Massa, Lerer, Bradbury, Chanan,
  Killeen, Lin, Gimelshein, Antiga, Desmaison, Kopf, Yang, DeVito, Raison,
  Tejani, Chilamkurthy, Steiner, Fang, Bai, and Chintala}]{paszke2015}
Paszke, A.; Gross, S.; Massa, F.; Lerer, A.; Bradbury, J.; Chanan, G.; Killeen,
  T.; Lin, Z.; Gimelshein, N.; Antiga, L.; Desmaison, A.; Kopf, A.; Yang, E.;
  DeVito, Z.; Raison, M.; Tejani, A.; Chilamkurthy, S.; Steiner, B.; Fang, L.;
  Bai, J.; and Chintala, S. 2019.
\newblock PyTorch: An Imperative Style, High-Performance Deep Learning Library.
\newblock In \emph{Advances in Neural Information Processing Systems
  (NeurIPS)}, 8024--8035.

\bibitem[{Puiutta and Veith(2020)}]{puiutta2020}
Puiutta, E.; and Veith, E. M. S.~P. 2020.
\newblock Explainable Reinforcement Learning: {A} Survey.
\newblock In \emph{Machine Learning and Knowledge Extraction International
  Cross-Domain Conference {(CD-MAKE)}}, volume 12279, 77--95.

\bibitem[{Ribeiro, Singh, and Guestrin(2016)}]{ribeiro2016}
Ribeiro, M.~T.; Singh, S.; and Guestrin, C. 2016.
\newblock ``Why Should {I} Trust You?": Explaining the Predictions of Any
  Classifier.
\newblock In \emph{{SIGKDD} International Conference on Knowledge Discovery and
  Data Mining}, 1135--1144. {ACM}.

\bibitem[{Sidrane et~al.(2021)Sidrane, Maleki, Irfan, and
  Kochenderfer}]{sidrane2021}
Sidrane, C.; Maleki, A.; Irfan, A.; and Kochenderfer, M.~J. 2021.
\newblock {OVERT:} An Algorithm for Safety Verification of Neural Network
  Control Policies for Nonlinear Systems.
\newblock \emph{Computing Research Repository}, abs/2108.01220.

\bibitem[{Sutton and Barto(2018)}]{sutton1998}
Sutton, R.~S.; and Barto, A.~G. 2018.
\newblock \emph{Reinforcement Learning: An Introduction}.
\newblock The MIT Press, second edition.

\bibitem[{van Hasselt(2010)}]{hasselt2010}
van Hasselt, H. 2010.
\newblock Double Q-learning.
\newblock In \emph{Advances in Neural Information Processing Systems
  (NeurIPS)}, 2613--2621.

\bibitem[{Verma et~al.(2018)Verma, Murali, Singh, Kohli, and
  Chaudhuri}]{verma2018}
Verma, A.; Murali, V.; Singh, R.; Kohli, P.; and Chaudhuri, S. 2018.
\newblock Programmatically Interpretable Reinforcement Learning.
\newblock In \emph{International Conference on Machine Learning (ICML)},
  volume~80, 5052--5061.

\end{thebibliography}

% \section{Acknowledgments}
% Placeholder % Anil Yildiz

\end{document}